\pdfoutput=1

\documentclass[11pt]{article}
\usepackage{blindtext} 
\makeatletter
\newcommand\blfootnote[1]{%
  \begingroup
  \renewcommand\thefootnote{\fnsymbol{footnote}}%
  \footnotetext{#1}%
  \endgroup
}
\makeatother

\usepackage[preprint]{acl}
\usepackage{algorithm}
\usepackage{times}
\usepackage{latexsym}

\usepackage[T1]{fontenc}

\usepackage[utf8]{inputenc}

\usepackage{microtype}

\usepackage{inconsolata}
\usepackage{amsmath}

\usepackage{graphicx}

\usepackage{microtype}
\usepackage{caption}
\usepackage{subcaption}
\usepackage{booktabs} 
\usepackage[utf8]{inputenc} 
\usepackage[T1]{fontenc}    
\usepackage{url}            
\usepackage{booktabs}       
\usepackage{amsfonts}       
\usepackage{nicefrac}       
\usepackage{microtype}      
\usepackage{xcolor}         
\usepackage{wrapfig}
\usepackage{booktabs}
\usepackage{enumitem}
\usepackage{makecell}
\usepackage{diagbox}
\usepackage{bm}
\usepackage{dsfont}
\usepackage{multirow}
\usepackage{varwidth}
\usepackage{pifont}
\usepackage{makecell}
\usepackage{wrapfig}
\usepackage{graphicx} 
\usepackage{comment}
\usepackage{color}
\usepackage[colaction]{multicol}
\usepackage{colortbl}
\usepackage{algpseudocode} 
\usepackage{xspace}
\usepackage{rotating}
\usepackage{longtable}
\usepackage{adjustbox}
\usepackage{threeparttable}
\usepackage{listings}

\usepackage{hyperref}
\definecolor{myblue}{HTML}{bfcdf0}
\definecolor{myred}{HTML}{e07f7f}

\newcommand{\aname}{\textsc{ITT}\xspace}

\title{Inner Thinking Transformer: \\ Leveraging Dynamic Depth Scaling to Foster Adaptive Internal Thinking}

\author{
    Yilong Chen$^{1,2}$,
    ~Junyuan Shang$^{3\ddagger}$,
    ~Zhenyu Zhang$^{3}$,
    ~Yanxi Xie$^{4}$,
    ~\textbf{Jiawei Sheng}$^1$,
    ~\textbf{Tingwen Liu}$^{1,2\dagger}$, 
    \\~\textbf{Shuohuan Wang}$^3$\textbf{,}~\textbf{Yu Sun}$^3$\textbf{,}~\textbf{Hua Wu}$^3$\textbf{,}~\textbf{Haifeng Wang}$^3$  \\ 
    \normalsize $^1$ Institute of Information Engineering, Chinese Academy of Sciences\\
    \normalsize $^2$ School of Cyber Security, University of Chinese Academy of Sciences\\
    \normalsize $^3$ Baidu Inc.\\
    \normalsize $^4$ School of Artificial Intelligence, Beijing Normal University\\
    \small \{\texttt{chenyilong, shengjiawei, liutingwen\}@iie.ac.cn} \\
    \small \{\texttt{shangjunyuan, zhangzhenyu07, wangshuohuan, sunyu02\}@baidu.com} \\
}

\begin{document}
\maketitle
\begin{abstract}

Large language models (LLMs) face inherent performance bottlenecks under parameter constraints, particularly in processing critical tokens that demand complex reasoning. Empirical analysis reveals challenging tokens induce abrupt gradient spikes across layers, exposing architectural stress points in standard Transformers. Building on this insight, we propose Inner Thinking Transformer (ITT), which reimagines layer computations as implicit thinking steps. ITT dynamically allocates computation through Adaptive Token Routing, iteratively refines representations via Residual Thinking Connections, and distinguishes reasoning phases using Thinking Step Encoding. ITT enables deeper processing of critical tokens without parameter expansion. Evaluations across 162M-466M parameter models show ITT achieves 96.5\% performance of a 466M Transformer using only 162M parameters, reduces training data by 43.2\%, and outperforms Transformer/Loop variants in 11 benchmarks. By enabling elastic computation allocation during inference, ITT balances performance and efficiency through architecture-aware optimization of implicit thinking pathways.

\end{abstract}

\blfootnote{$^\dagger$Corresponding author. $^\ddagger$ Project lead. Preliminary work.}

\section{Introduction}

Large language models (LLMs)~\cite{claude,Gpt-4,touvronLlamaOpenFoundation2023}  have demonstrated remarkable performance across numerous natural language tasks. Recent studies~\cite{fernandez2024hardware, hoffmann2022training, chen2024scaling} indicate that scaling laws for LLM parameters exhibit diminishing returns under constrained data availability and computational resource budgets. Scaling model parameters increases computational and deployment costs, making high-performance models impractical for resource-constrained environments. Meanwhile, smaller models encounter performance bottlenecks primarily attributable to limited parameter space.

Recent approaches, such as Test-Time Scaling ("Slow-Thinking")~\cite{muennighoff2025simple,snell2024scaling,ma2024inference}, aim to enhance performance by allocating more computation during the inference search process. While effective, these methods are limited by the reliance on accurately generating key tokens, which can lead to catastrophic reasoning failures~\cite{chen2023token,singh2024exposing,jiang2024peek}, especially in smaller models. Some works enhance model performance through layer sharing~\cite{li2024lisa,li2024crosslayer}, recursion~\cite{ng2024loopneuralnetworksparameter,dehghani2019universaltransformers,geiping2025scalingtesttimecomputelatent}, or implicit reasoning~\cite{deng2023implicit,shalev2024distributional}, but they fail to flexibly improve the model's reasoning ability on key tokens, which either suffer from insufficient performance or  redundant overhead.


\begin{figure}[t]  
\centering  
\includegraphics[width=7.6cm]{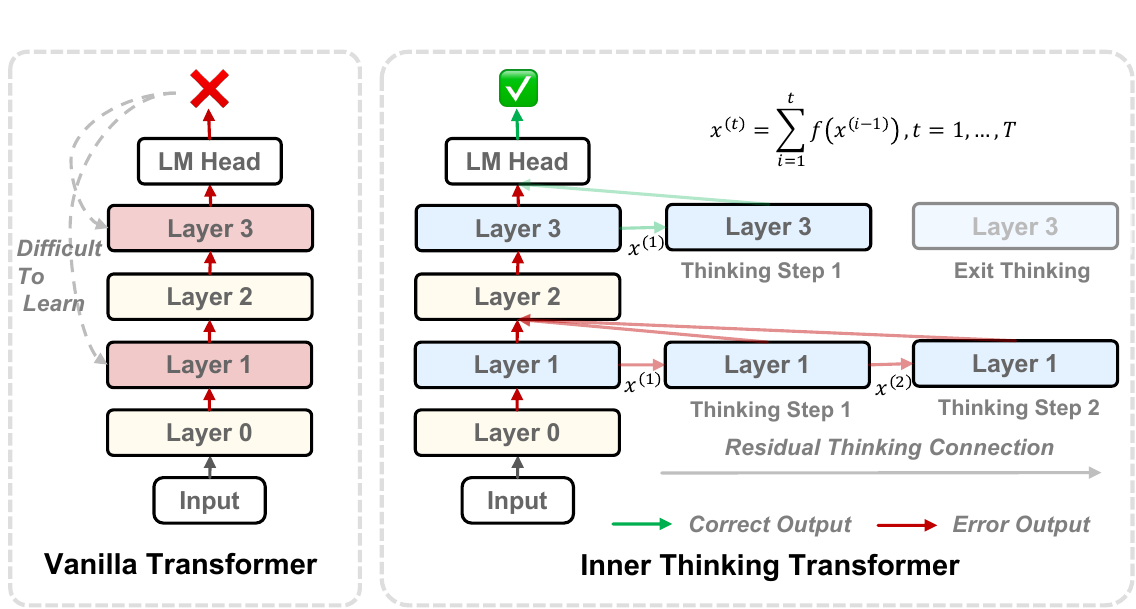}
\caption{
The Transformer, constrained by a limited number of parameters, tends to make errors on difficult samples. We treat each single computation in the model's layers as one step of inner thinking. By training the model to allocate more inner thinking steps at specific layers and organize thinking results, the model can achieve better results without scaling parameters.
}
\vspace{-2mm}
\label{fig:intro_motivation}
\end{figure}


In this work, we aim to explore\textit{ how the model can allocate more computation to individual tokens, enhancing testing performance without increasing parameters.}
Through analysis in Section~\ref{sec.obs}, we explore how models learn and reason about critical tokens. Our findings reveal that \textit{simple tokens are resolved efficiently in early layers with stable low-gradient flows, while complex tokens cause difficulties across layers, with sudden gradient spikes indicating architectural or parametric issues.} The differentiated properties of layers inspire us to propose a novel perspective on the model’s internal reasoning process: \textit{Inner Thinking}.  Inner thinking conceptualizes the evolution of hidden states layer by layer, with each layer representing a distinct implicit reasoning step for deriving a single token.

Intuitively, we can extend and combine multiple inner thinking steps to break the model's performance bottleneck.
Therefore, we propose a novel approach called Inner Thinking Transformer (\aname). \aname enhances token-level reasoning by dynamically allocating additional thinking steps to key tokens and iteratively accumulating residual thinking results to refine tokens' representations. As shown in Figure~\ref{fig:intro_motivation}, the model learns to “think” more deeply on important information during training. Specifically, we design a dynamic token-wise depth architecture based on\textit{ Adaptive Token Routing} networks and adopt a\textit{ Residual Thinking Connection} mechanism (RTC) that gradually converges toward better outcomes at each step. In addition, we introduce a\textit{ Thinking Step Encoding} scheme to better differentiate between successive thinking steps.

Notably, while trained under specific thinking settings, our architecture can \textit{flexibly allocate more computational resources during testing time} to improve performance or achieve a balanced trade-off between resources and performance (see Figure~\ref{fig:res_3figs}). \textit{The routing network autonomously develops a thinking pattern that strategically balances depth and breadth}: specific thinking steps are allocated for intensive processing of complex tokens, while more efficient pathways handle simpler tokens (see Figure~\ref{fig:router_visual}). In general, \aname mitigates the performance bottleneck in reasoning for individual tokens and can be combined with COT methods to resolving reasoning challenges for critical tokens.

Experimentally, we construct both vanilla Transformer,  Loop variants and \aname variants across three scales (162M, 230M, and 466M parameters) following the LLaMA architecture. Evaluated on an 11-task benchmark, \aname consistently outperforms Transformer and Loop variants with an equivalent parameters. \aname achieves higher performance with the same FLOPs and saves 43.2\% of the training data budget compared to Transformer. Notably, the ITT ×4 -162M model significantly surpasses the 230M Transformer and even achieves 96.5\% performance of 466M Transformer. Overall, \aname introduces an inherent test-time scaling in the model, achieving both performance and efficiency balance through its elastic deep computation paradigm.

\section{Observation} \label{sec.obs}
\begin{figure}[t]  
\centering  
\includegraphics[width=7.6cm]{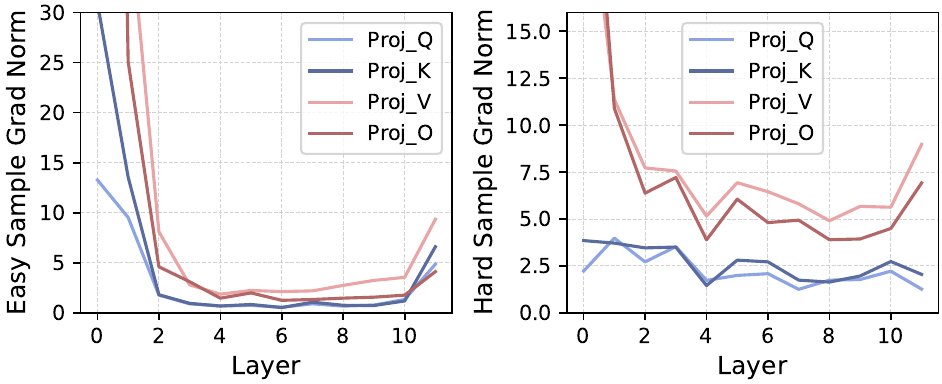}
\caption{
Layer's Gradient Nuclear Norm of the Attention matrices of GPT-2 on hard or simple samples.
}
\vspace{-2mm}
\label{fig:obs}
\end{figure}

To investigate how models learn about critical tokens, our empirical analysis of GPT-2's attention matrices through gradient nuclear norm (GNN) measurements~\cite{li2024happenedllmslayerstrained} reveals systematic patterns in layer-wise dynamics. Using the AQuA corpus~\cite{ling2017program}, we firstly train GPT-2 in 100 samples then categorize samples in evaluation into easy (model answers correctly) and hard (model answers incorrectly). In Figure~\ref{fig:obs}, for easy samples, GNN values decay exponentially across early layers (L0-L2) and final layers (L11), stabilizing below 3 in layers (L3-L10). In contrast, hard samples exhibit persistent GNN oscillations throughout all 12 layers, punctuated by abrupt spikes at strategic layer positions (L3, L5, L7, L9).

These observations reveal one of the underlying reasons for the presence of hard-to-learn samples in models: as shown in Figure~\ref{fig:intro_motivation}, certain parameters face significant optimization difficulties due to architectural limitations (e.g., insufficient depth) or parameter constraints. Many studies suggest that Transformer layers exhibit unique functional characteristics and training variances~\cite{alizadeh2024duollmframeworkstudyingadaptive,sun2025transformerlayerspainters,takase2023lessonsparametersharinglayers}.. This inspires us to propose a framework where \textit{each layer transformation in the model is viewed as a single thinking step on latent information.} By studying the inner thinking process, we aim to
design corresponding architectures to optimize model's learning difficulty and inference performance.

\section{Method} \label{sec.method}
\begin{figure*}[t]
\begin{minipage}[c]{\textwidth}
\centering
  \includegraphics[width=\textwidth]{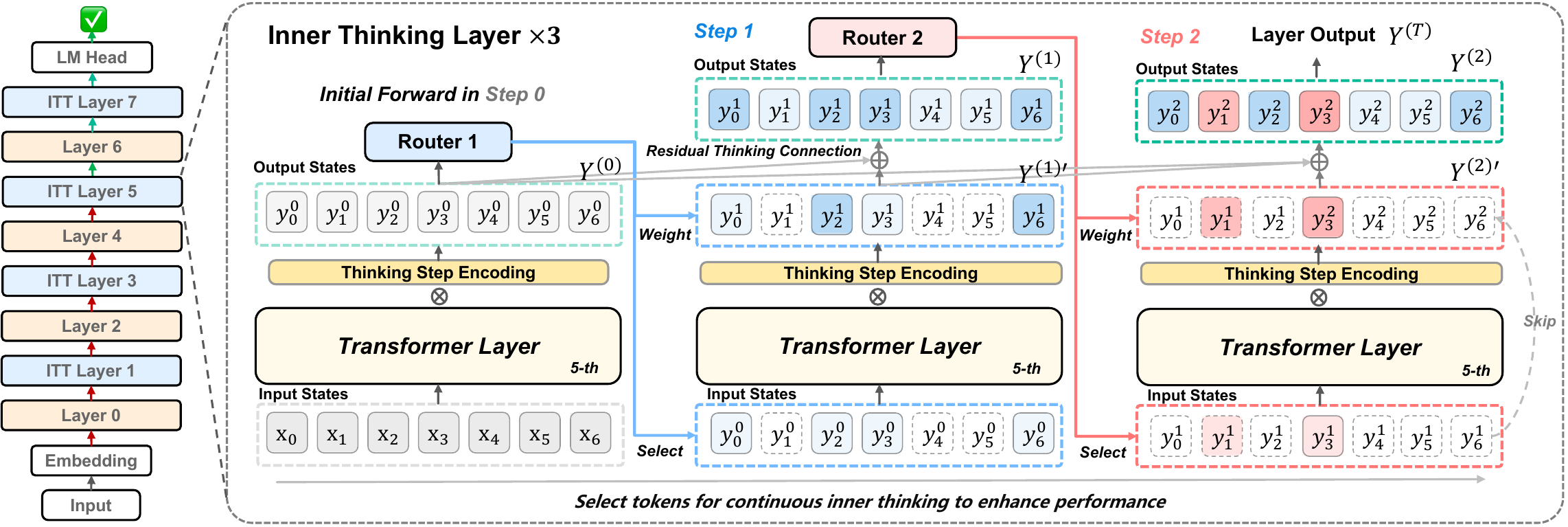}
  \caption{
  An illustration of \aname: ITT uses Adaptive Token Routing to select and weight important tokens for each inner thinking step. Based on Thinking Step Encoding and Residual Thinking Connection, ITT layer iterates thinking multiple times, accumulating each step's results for improved layer output.
  }\label{fig:main}
\end{minipage}
\vspace{-2mm}
\end{figure*}

In this section, we introduce our Inner Thinking (\aname) framework (Figure~\ref{fig:main}) to enhance transformer models by dynamically deepening token-level reasoning. We begin in Section~\ref{method:step} by formalizing inner thinking steps within the transformer. Section~\ref{method:residual} then details the Residual Thinking Connection, where inner steps are extended via residual accumulation. In Section~\ref{method:routing}, we present the Adaptive Token Routing, which employs a weight predictor to select the most critical tokens for further thinking. Finally, Section~\ref{method:bp} demonstrate how \aname enhances learning efficiency in backporpogation.

\subsection{Inner Thinking Step in Transformer}
\label{method:step}

Traditional reasoning in Transformer models typically relies on token-by-token generation. Given an input $x$, the output sequence $y = (y_1, y_2, \ldots, y_N)$ is generated as
\begin{equation} \small
P(y \mid x) = \prod_{n=1}^{N} P(y_n \mid y_{<n}, x),
\label{eq:sequentially} \end{equation}
However, errors in key tokens can propagate, potentially leading to an incorrect result. To investigate the intrinsic mechanisms in single-token generating, we propose a novel concept of \emph{Inner Thinking} in model's depth that decomposes the generation of each token into a series of internal thinking steps. Specifically, given an initial state $x^{(0)}$, we define Inner Thinking as
\begin{equation} \small
X^{(t)} = f^{(t)}\big(x^{(t-1)}\big), \quad t = 1, 2, \ldots, T,
\label{eq:inner_thinking} \end{equation}
where $f^{(t)}(\cdot)$ represents the transformation corresponding to the $t$-th thinking step (consist of one or more Transformer layers) and $T$ is the maximum number of steps. The final token is then generated based on the output of the last thinking step:
\begin{equation} \small
P(y \mid x) = \operatorname{softmax}\big(W\, x^{(T)} + b\big),
\label{eq:lmout} \end{equation}
with $W$ and $b$ denoting the weights and bias for the output projection. Define $\mathcal{L}(\cdot, y)$ measures the discrepancy between final state $X^{(T)}$ and the target token $y$, we have two scenarios:

\paragraph{Early Exit:} If at an intermediate step $t_0 < T$, the state $x^{(t_0)}$ is close enough to the target (i.e., $\mathcal{L}\big(x^{(t_0)}, y\big) < \epsilon$, where $\epsilon$ is a threshold), the model can stop and output the token as $y = \psi\big(x^{(t_0)}\big)$, where $\psi(\cdot)$ is the decoding function. This allows the model to achieve correct results with fewer Inner Thinking Steps, improving efficiency.
    
\paragraph{Performance Deficiency:} Conversely, if even after all $T$ internal steps the discrepancy remains high (i.e., $\mathcal{L}\big(x^{(T)}, y\big) > \epsilon$), it indicates that the Inner Thinking was insufficient to correctly approximate the target. This scenario highlights potential areas for improvement in the model's reasoning capacity or its internal step design.

\subsection{Residual Thinking Connection}
\label{method:residual}

Under the framework defined in Section~\ref{method:step}, we aim to enhance the model's performance to reduce \textbf{Performance Deficiencies}. For challenging examples, high gradient values are observed in Section~\ref{sec.obs}, indicating that the model faces optimization difficulties. 
To address these issues, a natural approach is to increase the number of inner thinking steps in one layer's computation. Therefore, we propose a \emph{Residual Thinking Connection} (RTC) mechanism that train model's layer parameters to\textbf{ learn iterative thinking capabilities}, reducing the difficulty of single-step thinking and enabling multiple uses of parameters to break performance bottlenecks.

Let $x^{(0)} \in \mathbb{R}^{d}$ denote RTC Layer input of a token representation, where $d$ is the hidden dimension. We denote $f:\mathbb{R}^{d}\rightarrow\mathbb{R}^{d}$ as the layer  transformation,  $T$ is the maximum number of thinking steps. In RTC, the final output after $t$ iterative steps is computed by cumulatively accumulating each step's outputs:
\begin{equation}
\small
\begin{split}
x^{(t)} &= \sum_{i=1}^{t} \left( f\big(x^{(i-1)}\big) \odot \phi^{(i)}  \right), 
t = 1, \ldots, T \\
\end{split}
\label{eq:rtc}
\end{equation}
where $\phi^{(t)} \in \mathbb{R}^{d}$ the learnable thinking position encoding associated with the $t$-th inner thinking step, which measuring the differences and importance of each step. Rather than processing the input representation only once, RTC Layer iteratively refine it by adding the residual contributions of each step's layer-output together with a learnable encoding. 
Compared to direct looping~\cite{ng2024loopneuralnetworksparameter,dehghani2019universaltransformers},\textit{ RTC not only enables deeper thinking but also effectively measures and combines each thinking step, allowing them to complement each other.} RTC provides the foundation for scaling Inner Thinking during testing.

\subsection{Adaptive Token Routing}
\label{method:routing}

 RTC in Section~\ref{method:residual} provides a method to enhance inner thinking. However,\textit{ different tokens require a varying number of thinking steps in the model}, as show in Section~\ref{sec.obs}. Moreover, we aim for the model to learn detailed, task-specific information at each step. To avoid unnecessary computation and information interference from processing all tokens at once, we introduce Adaptive Token Routing (ATR). Inspired by deep conditional computation~\cite{raposo2024mixtureofdepthsdynamicallyallocatingcompute,zhang2024pmodbuildingmixtureofdepthsmllms}, ATR, based on a routing network, selects the most important tokens for thinking at each step.
 
 Let the input sequence be denoted by \( X \in \mathbb{R}^{n \times d} \), where \( n \) is the sequence length. We perform a forward pass to obtain the output \( Y^{(0)} \in \mathbb{R}^{n \times d} \), and then linear weight predictor $\mathcal{R}^{(0)} \in \mathbb{R}^{d \times 1} $ is applied to \( Y^{(0)} \) to generate an importance score:
 \begin{equation} \small
Y^{(0)} = f(X), \quad
w^{(1)} = \mathcal{R}^{(1)}(Y^{(0)}) \in \mathbb{R}^{n},
\end{equation}
 and we denote by \(P_\rho(w^{(1)})\) the \(\rho\)-th percentile of these scores, with \(\rho\) being a predefined selection ratio. For a given thinking step \(t\), the calculation process in \aname layer can be formulated as:
\begin{equation} \small
Y_i^{(t)\prime}=\left\{\begin{array}{lll}
\alpha^{(t)} w^{(t)}_i f\left(Y^{(t-1)}_i\right), & \text { if } \quad w^{(t)}_i>P_\rho(w^{(t)}), \\
Y^{(t-1)}_i, & \text { if } \quad w^{(t)}_i \leq P_\rho(w^{(t)}),
\end{array}\right.
\end{equation}
where $\alpha^{(t)}$ is a hyperparam in $t$ step, \( w^{(t)}_i>P_\rho(w^{(t)})\) is the indicator function selecting only the tokens with predicted weights exceeding the threshold. The router \(\mathcal{R}^{(t)}\)\textit{modulates the decision to execute an additional thinking iteration based on the current token representation and the step-specific encoding.} For tokens deemed important, the model applies an extra weighted transformation. Conversely, tokens that do not meet the selection criteria bypass the extra processing, preserving their previous representation. The router's weights are part of the gradient path, allowing the routing parameters to be updated through backpropagation. 

Finally, \aname (in Figure~\ref{fig:main}) combine the results of each step using RTC, following Equation~\ref{eq:rtc}:
\begin{equation}
\small
\begin{split}
Y^{(t)} &= Y^{(0)} \odot \phi^{(0)}  + \sum_{i=1}^{t} \left( Y_i^{(i)\prime} \odot \phi^{(i)}  \right) , \\
t &= 1, \ldots, T.
\end{split}
\end{equation}
This unified update thus integrates RTC with dynamic, token-level routing, enabling the model to \textit{adaptively allocate computational resources only where deeper thinking is required}. By iteratively selecting a subset of tokens for deeper processing, the model can efficiently reinforce key tokens without increasing the model parameter.
In practice, the ITT layer can be flexibly improved based on the model layers. We insert the ITT layer at regular intervals alongside the model's original layers to construct a flexible inner thinking model, and optimize all model parameters using the language modeling cross-entropy loss: $\mathbb{L} = \mathbb{L}_{\text{CE}}$.

\subsection{Optimization}
\label{method:bp}

In this section, we prove Residual Thinking Learning \textit{extends single-step optimization into multi-step optimization,  making it easier to converge} during backpropagation compared to a direct one-step mapping. Let \( y^* \in \mathbb{R}^d \) the corresponding ground-truth, \( \Theta' \) represents the origin Layer parameters, and \( \theta \) represents th \aname layer parameters. The optimization objective is to minimize the loss:
\begin{equation} \small
\mathcal{L}\bigl(F(x; \Theta', \theta), y^*\bigr) = \mathcal{L}\bigl(G(f_T(x; \theta); \Theta'), y^*\bigr).
\end{equation}
For each step's parameter \( \theta \), the gradient is computed using the chain rule:
\begin{equation} \small
\frac{\partial \mathcal{L}}{\partial \theta} = \frac{\partial \mathcal{L}}{\partial Y^{(t)}} \cdot \prod_{j=t+1}^{T} \left[I + \frac{\partial \Delta_j(Y^{(j)}; \theta)}{\partial Y^{(j)}}\right] \cdot \frac{\partial \Delta_k(Y^{(0)}; \theta)}{\partial \theta}.
\end{equation}
Assuming that the corrections \( \Delta_j \) are small, we can approximate the product term by the identity matrix \( I \), yielding:
\begin{equation} \small
\frac{\partial \mathcal{L}}{\partial \theta} \approx \frac{\partial \mathcal{L}}{\partial Y^{(t)}} \cdot \frac{\partial \Delta_k(Y^{(0)}; \theta)}{\partial \theta}.
\end{equation}
This shows that the gradient update at each small step is nearly equal to the global gradient multiplied by the derivative of the local mapping, aligning with global loss reduction. Assuming each iteration reduces the error by a factor of \( c \), this leads to exponential decay \( c^t \), proving that iterative corrections ensure stable, efficient convergence. In summary, our method avoids excessive scaling or distortion from deep chain propagation. It extends single-step optimization to multi-step, easing convergence and preventing gradient vanishing or explosion.

\section{Experiments}

\begin{table*}[!ht]
\setlength\tabcolsep{5.5pt}
\footnotesize
    \centering
    \small
    \begin{tabular*}{1.0\textwidth}{@{\extracolsep{\fill}}@{}l c cccccc cc cc c @{}}
    \toprule
    \multirow{2}{*}{\raisebox{-0.5\height}{\textbf{Model-Params}}} & \multirow{2}{*}{\raisebox{-0.5\height}{\textbf{FLOPs}}} & \multicolumn{6}{c}{\scriptsize\textbf{Commonsense \& Reading Comprehension}} & \multicolumn{2}{c}{\scriptsize\textbf{Continued}} &  \multicolumn{1}{c}{\scriptsize\textbf{LM}} & \multicolumn{1}{c}{\scriptsize\textbf{Knowledge}}  & \multirow{2}{*}{\raisebox{-0.5\height}{\textbf{Avg.}}} \\
    \cmidrule(lr){3-8} \cmidrule(lr){9-10} \cmidrule(lr){11-11} \cmidrule(lr){12-12}
    & & \scriptsize\textbf{SciQ} & \scriptsize\textbf{PIQA} & \scriptsize\textbf{WG} & \scriptsize\textbf{ARC-E } & \scriptsize\textbf{ARC-C} & \scriptsize\textbf{Hella.} & \scriptsize\textbf{LogiQA}  & \scriptsize\textbf{BoolQ} & \scriptsize\textbf{Lam.} & \scriptsize\textbf{MMLU} \\
    \midrule
    LLaMA2-162M & 1.88 & 72.0 & 62.7 & 51.9 & 41.7 & 19.2 & 28.8 & 24.0 & 50.3 & 28.6 & 25.2 & 40.4 \\
    \midrule
     \rowcolor{myblue!50} Loop~$\times3$-162M & 3.76  & \textcolor{red!60!black}{71.8} & \textcolor{green!60!black}{63.1} & \textcolor{green!60!black}{53.0} & \textcolor{red!60!black}{40.4} & \textcolor{red!60!black}{19.1} & \textcolor{green!60!black}{29.1} & \textcolor{red!60!black}{20.9} & \textcolor{green!60!black}{51.9} & \textcolor{green!60!black}{28.8} & \textcolor{green!60!black}{25.7} & 40.4 \\
     \rowcolor{myblue!50}   Loop~$\times4$-162M & 4.70 & \textcolor{green!60!black}{72.8} & \textcolor{red!60!black}{62.4} & \textcolor{green!60!black}{52.6} & \textcolor{green!60!black}{41.8} & \textcolor{green!60!black}{19.8} & \textcolor{green!60!black}{29.4} & \textcolor{red!60!black}{22.0} & \textcolor{red!60!black}{49.9} & \textcolor{green!60!black}{30.1} & \textcolor{green!60!black}{26.3} & \textcolor{green!60!black}{40.7} \\
    \rowcolor{myred!15}    \aname$\times2$ -162M & 2.72 & \textcolor{green!60!black}{72.1} & \textcolor{green!60!black}{63.5} & \textcolor{green!60!black}{52.1} & \textcolor{red!60!black}{41.1} & \textcolor{red!60!black}{19.2} & \textcolor{green!60!black}{29.1} & \textcolor{red!60!black}{21.4} & \textcolor{green!60!black}{51.4} & \textcolor{green!60!black}{29.2} & \textcolor{green!60!black}{25.5} & \textcolor{green!60!black}{40.6} \\
    \rowcolor{myred!15}    \aname$\times3$ -162M & 3.19 & \textcolor{green!60!black}{73.9} & \textcolor{red!60!black}{62.5} & \textcolor{red!60!black}{50.6} & \textcolor{green!60!black}{43.6} & \textcolor{green!60!black}{19.3} & \textcolor{green!60!black}{29.2} & \textcolor{red!60!black}{20.6} & \textcolor{green!60!black}{52.1} & \textcolor{green!60!black}{37.1} & \textcolor{green!60!black}{25.8} & \textcolor{green!60!black}{41.5} \\
    \rowcolor{myred!15}    \aname$\times4$ -162M & 3.29 & \textcolor{green!60!black}{72.4} & \textcolor{green!60!black}{63.9} & \textcolor{green!60!black}{52.3} & \textcolor{green!60!black}{43.4} & \textcolor{green!60!black}{20.5} & \textcolor{green!60!black}{29.3} & \textcolor{green!60!black}{22.8} & \textcolor{green!60!black}{56.8} & \textcolor{green!60!black}{33.9} & \textcolor{green!60!black}{26.0} & \textcolor{green!60!black}{42.1} \\

    \midrule
    LLaMA2-230M & 2.87 &  72.8 & 65.0 & 49.3 & 44.0 & 19.9 & 29.1 & 20.6 & 60.2 & 31.7 & 25.5 & 41.8 \\
    \midrule
     \rowcolor{myblue!50} Loop~$\times3$-230M & 3.59 & \textcolor{red!60!black}{71.1} & \textcolor{red!60!black}{64.3} & \textcolor{green!60!black}{51.5} & \textcolor{red!60!black}{41.7} & \textcolor{green!60!black}{20.3} & \textcolor{green!60!black}{30.2} & \textcolor{green!60!black}{22.6} & \textcolor{green!60!black}{61.2} & \textcolor{green!60!black}{33.5} & \textcolor{green!60!black}{26.4} & \textcolor{green!60!black}{42.3} \\
        \rowcolor{myblue!50} Loop~$\times4$-230M & 3.95 & \textcolor{green!60!black}{74.1} & \textcolor{green!60!black}{65.1} & \textcolor{green!60!black}{52.0} & \textcolor{red!60!black}{41.7} & \textcolor{green!60!black}{20.1} & \textcolor{green!60!black}{30.2} & \textcolor{red!60!black}{18.6} & \textcolor{green!60!black}{61.0} & \textcolor{green!60!black}{32.5} & \textcolor{green!60!black}{26.7} & \textcolor{green!60!black}{42.2} \\

        \rowcolor{myred!15} \aname$\times2$ -230M & 3.19 & \textcolor{red!60!black}{72.7} & \textcolor{red!60!black}{64.6} & \textcolor{green!60!black}{52.2} & \textcolor{red!60!black}{43.3} & \textcolor{green!60!black}{20.5} & \textcolor{green!60!black}{29.7} & \textcolor{green!60!black}{22.0} & \textcolor{red!60!black}{59.7} & \textcolor{green!60!black}{32.6} & \textcolor{green!60!black}{25.9} & \textcolor{green!60!black}{42.3} \\
        \rowcolor{myred!15} \aname$\times3$ -230M & 3.37 & \textcolor{green!60!black}{74.3} & \textcolor{green!60!black}{65.7} & \textcolor{green!60!black}{52.8} & \textcolor{green!60!black}{44.9} & \textcolor{green!60!black}{20.8} & \textcolor{green!60!black}{30.8} & \textcolor{green!60!black}{23.1} & \textcolor{green!60!black}{62.5} & \textcolor{green!60!black}{34.2} & \textcolor{green!60!black}{26.3} & \textcolor{green!60!black}{43.5} \\
         \rowcolor{myred!15} \aname$\times4$ -230M & 3.41 & \textcolor{green!60!black}{75.1} & \textcolor{green!60!black}{66.2} & \textcolor{green!60!black}{53.5} & \textcolor{green!60!black}{45.0} & \textcolor{green!60!black}{21.1} & \textcolor{green!60!black}{31.2} & \textcolor{green!60!black}{22.4} & \textcolor{green!60!black}{62.7} & \textcolor{green!60!black}{34.8} & \textcolor{green!60!black}{26.6} & \textcolor{green!60!black}{43.9} \\
    \midrule
    LLaMA2-466M & 4.92 & 75.5 & 66.5 & 51.5 & 45.2 & 20.4 & 31.3 & 21.2 & 62.6 & 36.6 & 25.4 & 43.6 \\
    \midrule
        \rowcolor{myblue!50} Loop~$\times3$-466M & 6.15 & \textcolor{red!60!black}{74.3} & \textcolor{red!60!black}{65.8} & \textcolor{green!60!black}{52.9} & \textcolor{red!60!black}{44.0} & \textcolor{green!60!black}{21.0} & \textcolor{green!60!black}{32.0} & \textcolor{green!60!black}{22.5} & \textcolor{red!60!black}{59.2} & \textcolor{green!60!black}{37.2} & \textcolor{green!60!black}{26.1} & \textcolor{red!60!black}{43.5} \\
        \rowcolor{myblue!50} Loop~$\times4$-466M & 6.77 & \textcolor{green!60!black}{76.8} & \textcolor{green!60!black}{67.0} & \textcolor{red!60!black}{50.7} & \textcolor{green!60!black}{46.5} & \textcolor{green!60!black}{20.9} & \textcolor{green!60!black}{32.2} & \textcolor{red!60!black}{20.1} & \textcolor{red!60!black}{59.0} & \textcolor{green!60!black}{40.1} & \textcolor{red!60!black}{24.8} & \textcolor{green!60!black}{43.8} \\
        \rowcolor{myred!15} \aname$\times2$ -466M & 5.47 & \textcolor{green!60!black}{75.9} & \textcolor{red!60!black}{66.2} & \textcolor{green!60!black}{52.7} & \textcolor{green!60!black}{45.4} & \textcolor{green!60!black}{21.2} & \textcolor{green!60!black}{32.1} & \textcolor{green!60!black}{21.8} & \textcolor{red!60!black}{60.7} & \textcolor{green!60!black}{38.4} & \textcolor{green!60!black}{25.7} &  \textcolor{green!60!black}{43.9} \\
        \rowcolor{myred!15} \aname$\times3$ -466M & 5.78 & \textcolor{green!60!black}{77.9} & \textcolor{red!60!black}{66.4} & \textcolor{green!60!black}{53.7} & \textcolor{green!60!black}{46.7} & \textcolor{green!60!black}{22.0} & \textcolor{green!60!black}{32.8} & \textcolor{green!60!black}{22.6} & \textcolor{red!60!black}{59.1} & \textcolor{green!60!black}{39.3} & \textcolor{green!60!black}{26.7} & \textcolor{green!60!black}{44.7} \\
        \rowcolor{myred!15} \aname$\times4$ -466M & 5.84 & \textcolor{green!60!black}{77.2} & \textcolor{green!60!black}{67.1} & \textcolor{green!60!black}{54.3} & \textcolor{green!60!black}{47.3} & \textcolor{green!60!black}{22.4} & \textcolor{green!60!black}{32.3} & \textcolor{green!60!black}{22.7} & \textcolor{red!60!black}{61.9} & \textcolor{green!60!black}{40.8} & \textcolor{green!60!black}{27.0} & \textcolor{green!60!black}{45.3} \\
    \bottomrule
    \end{tabular*}
    \caption{Comprehensively evaluate the basic capabilities of models with different activated parameters. In particular, \aname$\times4$-162M represents a model with 162M total parameters using \aname to think total 4 steps. 
    }
    \label{table:main-results}
    \vspace{-2mm}
\end{table*}

\subsection{Setup}

\paragraph{Data.} To pretrain \aname models and baseline models, we employ the RedPajama~\cite{Redpajama}, which parallels the LLaMA training data across seven domains: CommonCrawl, C4, GitHub, Wikipedia, Books, ArXiv, and Stack-Exchange. This dataset comprises a 2 million tokens validation set and a 50 billion tokens training set.

\paragraph{Training.} Our experimental framework utilizes the Sheared-LLaMA codebase \cite{xiaShearedLLaMAAccelerating2023} implemented on the Composer package \cite{mosaicml2022composer}, and is executed on 8 NVIDIA A100 GPUs (80GB). The models are trained with a sequence length of 4096, employing a global batch size of 256.
\aname models are trained for 50000 steps (50B token budget). The learning rates were set at 3e-4 for all parameters. The baselines and all \aname models follow the same training setup, starting from random initialization and training on the same dataset.

\paragraph{Evaluation.} We employed the lm-evaluation-harness \cite{eval-harness} to evaluate our models. For common sense and reading comprehension tasks, we report 0-shot accuracy for SciQ \cite{sciqa}, PIQA \cite{piqa}, WinoGrande (WG) \cite{WinoGrande:conf/aaai/SakaguchiBBC20}, ARC Easy(ARC-E) \cite{clark2018think}, and 10-shot HellaSwag (Hella.) \cite{HellaSwag:conf/acl/ZellersHBFC19}, alongside 25-shot accuracy for ARC Challenge (ARC-C) \cite{arcChallenge:journals/corr/abs-1803-05457}. For continued QA and text understanding, we report 0-shot accuracy for LogiQA \cite{liu2020logiqa}, 32-shot BoolQ \cite{clark2019boolq}, and 0-shot LAMBADA (Lam.) \cite{paperno2016lambada}. All reported results are calculated with the mean and stderr of multiple experiments.

\paragraph{Baseline.} Following the architecture of LLaMA2, we constructed models at three parameter scales: 162M, 230M, and 466M, with hidden dimensions of 1024, 1536, and 2048, as shown in Table~\ref{tab:model_setting}. 
For each parameter scale, we develop three variants: 
\begin{itemize} 
\item Vanilla Transformers in LLaMA architecture~\cite{touvronLlamaOpenFoundation2023}. \vspace{-2mm}
\item The Loop Neural Network design~\cite{ng2024loopneuralnetworksparameter,dehghani2019universaltransformers} implements recurrence for iterative refinement. \vspace{-2mm}
\item Our \aname architecture, adaptively selecting a subset of tokens for deeper thinking.
\end{itemize}
We experiment with three thinking step scaling factors—$2\times$, $3\times$ and $4\times$.
We replace every other layer of original model with a Loop or \aname layer.

\subsection{Result}

\begin{figure*}[t]
\begin{minipage}[c]{\textwidth}
\centering
  \includegraphics[width=\textwidth]{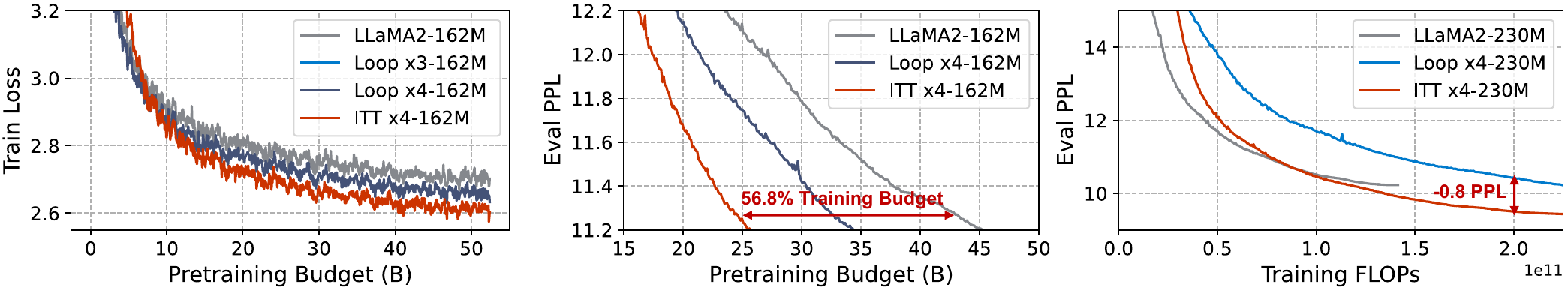}
  \caption{ \textbf{Left:} Loss curves for 162M-models pre-trained on 50B tokens. \textbf{Middle:} Eval Perplexity curves for 162M-models pre-trained on 50B tokens. \textbf{Right:} Eval Perplexity for 230M-models with Training FLOPs.}
  \label{fig:exp_3figs}
\end{minipage}
\vspace{-2mm}
\end{figure*}
\begin{figure}[t]  
\vspace{-2mm}
\centering  
\includegraphics[width=6cm]{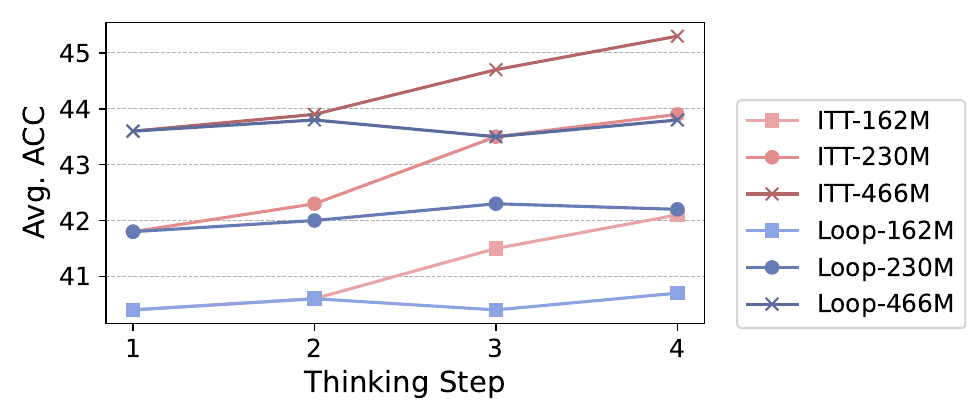}
\vspace{-3mm}
\caption{
Average accuracy after training 50B tokens for the ITT and Loop models (162M, 230M, 460M) under different thinking step configurations.
}
\vspace{-3mm}
\label{fig:elastic_scaling}
\end{figure}

\begin{figure*}[t]
\begin{minipage}[c]{\textwidth}
\centering
  \includegraphics[width=\textwidth]{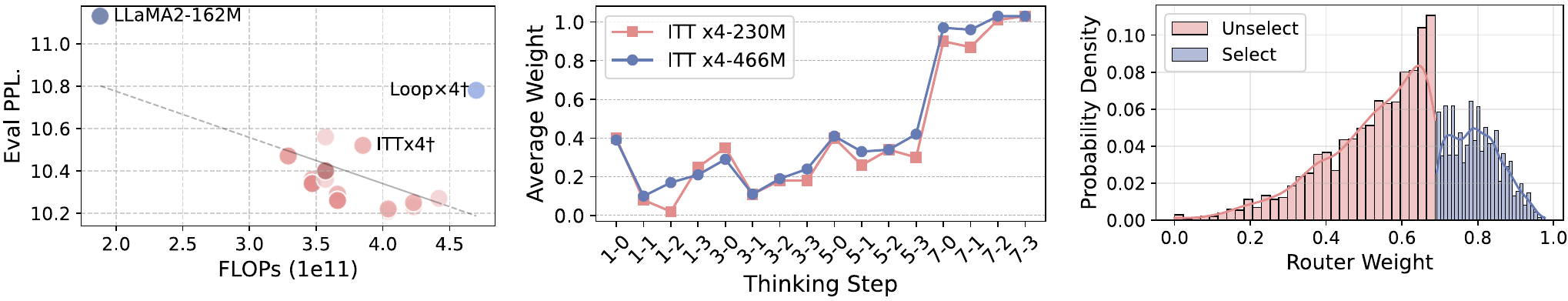}
  \caption{ \textbf{Left:} 
  Perplexity vs. FLOPs for different selection strategies. Lower left region indicates better performance-efficiency balance.
  \textbf{Middle:} The average weights by the learned Thinking Step Encoding in the ITT x4 model (230M, 460M) across different thinking steps.
  \textbf{Right:} 3-2 step's Router Weight Distribution in ITT $\times$4.}\label{fig:res_3figs}
\end{minipage}
\vspace{-2mm}
\end{figure*}

\paragraph{Foundational Capabilities.}
Table~\ref{table:main-results} shows the performance improvements of \aname (pink) and Loop (blue) on LLaMA 2's 162M, 230M, and 466M versions. Both methods enhance model performance by increasing computational allocation during training and inference without expanding parameters. \textbf{Thanks to its unique RTC design, \aname achieves better test-time scaling performance} than Loop, as shown in Figure~\ref{fig:elastic_scaling}. For example, the 162M \aname $\times$4 configuration improves the baseline by 1.7\% with 4-step deep thinking in 50\% of layers, while Loop improves only by 0.3\% after 4 iterations. \textbf{The advantages of \aname become clearer as model scale increases}, with improvements of 1.7\%, 2.1\%, and 1.7\% for the 162M, 230M, and 466M models. \aname shows overall enhancement across nearly all metrics, with notable improvements in ARC-E, BoolQ, and LAMBADA, reflecting gains in generative and reasoning abilities.
\paragraph{Convergence.} Figure~\ref{fig:exp_3figs} Left and Middle visualize the training loss and eval perplexity during 50B-token pre-training for LLaMA 2-2162M, Loop$\times$4, and \aname$\times$4. \textbf{\aname demonstrates superior training stability and efficiency}, with smoother, lower perplexity trajectories compared to LLaMA 2-230M and Loop. Notably, \aname$\times$4 shows a 0.09 loss reduction compared to baseline and 0.4 to Loop at 50B tokens. \textbf{\aname also reveals remarkable data efficiency}: it matches LLaMA 2-162M's performance using only 56.8\% of the training data, showcasing its capability in parameter-efficient scaling and data-efficient learning.

\paragraph{Computational Efficiency.} As shown in Figure~\ref{fig:exp_3figs} (Right), Figure~\ref{fig:res_3figs} (Left), and Table~\ref{table:main-results}, \textbf{\aname maintains high computational efficiency during test-time scaling}. With 3-step deep thinking, \aname incurs only 84\% of Loop's computational cost, dropping to 70\% at 4 steps. Remarkably, \textbf{\aname outperforms Loop with fewer computational FLOPs}, achieving performance similar to models with more parameters. Our experiments show that \aname$\times$2 outperforms Loop$\times$3 while using only 72\% of the computation and exceeds the 230M Dense model with just 70.4\% of the parameters. These results highlight the substantial computational efficiency gains from token-wise selective inner thinking in the \aname framework.

\paragraph{Elastic Thinking.} Our experiments show that \textbf{\aname models can elastically allocate computations for inner thinking.} As seen in Table~\ref{tab:elastic_main}, with 4-step thinking and 70\% token participation during training, we can flexibly adjust token selections to enhance performance (e.g., 10.21 PPL in the 70\%, 70\%, 90\% setting, 0.31 PPL lower than the training config), or reduce token selections to lower costs with no performance loss (e.g., 10.47 PPL in the 50\%, 50\%, 50\% setting). We can even \textit{remove a thinking step while maintaining near-identical results} to the training configuration. Figure~\ref{fig:res_3figs} Left shows the FLOPs and Eval PPL of ITT's elastic inference. Compared to the baselines, ITT achieves a performance-efficiency balance, with the dashed line illustrating \textbf{the near-linear tradeoff trend of ITT during testing}. ITT’s elastic thinking enables flexible deployment in diverse scenarios.

\begin{table}[t]
\centering
\small
\setlength{\tabcolsep}{2.5mm}{
\begin{tabular}{@{}lcc@{}}
\toprule
\textbf{Method - Select Ratio in Steps} & \textbf{FLOPs} & \textbf{Perplexity} $\downarrow$\\
\midrule
LLaMA2-162M & 1.88 & 11.13 \\
\midrule
\aname×4 - 90\%,~90\%,~90\% & 4.42 & 10.27 (\textcolor{green!70!black}{-0.86}) \\
\aname×4 - 90\%,~90\%,~0\% & 3.57 & 10.40 (\textcolor{green!70!black}{-0.73}) \\
\aname×4 - 90\%,~0\%,~90\% & 3.57 & 10.36 (\textcolor{green!70!black}{-0.77}) \\
\aname×4 - 0\%,~90\%,~90\% & 3.57 & 10.56 (\textcolor{green!70!black}{-0.57}) \\
\aname×4 - 90\%,~70\%,~90\% & 4.23 & 10.23 (\textcolor{green!70!black}{-0.90}) \\
\aname×4 - 70\%,~70\%,~90\% & 4.04 & 10.21 (\textcolor{green!70!black}{-0.92}) \\
\midrule
\aname×4 - 70\%,~70\%,~70\%$^\dagger$  & 3.85 & 10.52 (\textcolor{green!70!black}{-0.61}) \\
\midrule
\aname×4 - 70\%,~70\%,~50\% & 3.66 & 10.26 (\textcolor{green!70!black}{-0.87}) \\
\aname×4 - 70\%,~50\%,~50\% & 3.47 & 10.34 (\textcolor{green!70!black}{-0.79}) \\
\aname×4 - 50\%,~50\%,~50\% & 3.29 & 10.47 (\textcolor{green!70!black}{-0.66}) \\
\midrule
Loop×4 - 100\%,~100\%,~100\%$^\dagger$  & 4.70 & 10.78 (\textcolor{green!70!black}{-0.35}) \\
\bottomrule
\end{tabular}
}

\caption{Eval Perplexity with different token selection ratios for extended 3-steps thinking. $^\dagger$ refers to the model's training configuration.}\label{tab:elastic_main}
\vspace{-2mm}
\end{table}

\begin{table}[t]
\centering
\small
\setlength{\tabcolsep}{1.5mm}
\begin{tabular}{@{}llc@{}}
\toprule
\textbf{Method} & \textbf{FLOPs} & \textbf{Perplexity} $\downarrow$ \\
\midrule
\aname×4 -162M & 3.29 & 10.25 \\
\midrule
w/o Residual Thinking Connection & 3.29 & 11.02 (\textcolor{red!50!black}{+0.77}) \\
w/o Adaptive Token Routing & 4.70 & 10.44 (\textcolor{red!50!black}{+0.19}) \\
w/o Thinking Position Encoding & 3.29 & 10.56 (\textcolor{red!50!black}{+0.22}) \\
\midrule
Router Sampling (Top-K) & 3.29 & 10.25 (~~~~-~~~~) \\
Router Sampling (Top-P) & 3.29 & 10.34 (\textcolor{red!50!black}{+0.09}) \\
Router Weight Norm (Sigmoid) & 3.29 & 10.25 (~~~~-~~~~) \\
Router Weight Norm (Tanh) & 3.29 & 10.38 (\textcolor{red!50!black}{+0.13}) \\
Token Reweighting (Only Select) & 3.29 & 10.25 (~~~~-~~~~) \\
Token Reweighting (Symmetric) & 3.29 & 10.41 (\textcolor{red!50!black}{+0.16}) \\
\midrule
LLaMA2-162M & 1.88 & 11.13 (\textcolor{red!50!black}{+1.36}) \\
\bottomrule
\end{tabular}
\caption{Eval Perplexity with ablation on \aname×4 -162M. "w.o." indicates the method was ablated.}
\label{tab:abla}
\vspace{-2mm}
\end{table}

\subsection{Ablation Studies}

In Table~\ref{tab:abla}, we compare the ablation results of \aname×4 with 162M parameters to the baseline under zero-shot pretraining on 50B tokens, based on Eval PPL. The specific analysis is as follows:

\paragraph{Residual Thinking Connection.} Removing this core mechanism causes the largest performance drop (+0.77 PPL), validating our hypothesis about multi-step reasoning. The residual accumulation enables iterative refinement of token representations, particularly crucial for processing linguistically complex patterns. Without RTC, the model may also lose the ability for elastic computation.

\paragraph{Thinking Position Encoding.} Thinking Position Encoding provides the model with key information for each thinking step. As shown in Table~\ref{tab:abla}, removing it results in +0.31 PPL., as  model loses  information about importance of each thinking step. 

\begin{figure*}[t]
\begin{minipage}[c]{\textwidth}
\centering
  \includegraphics[width=\textwidth]{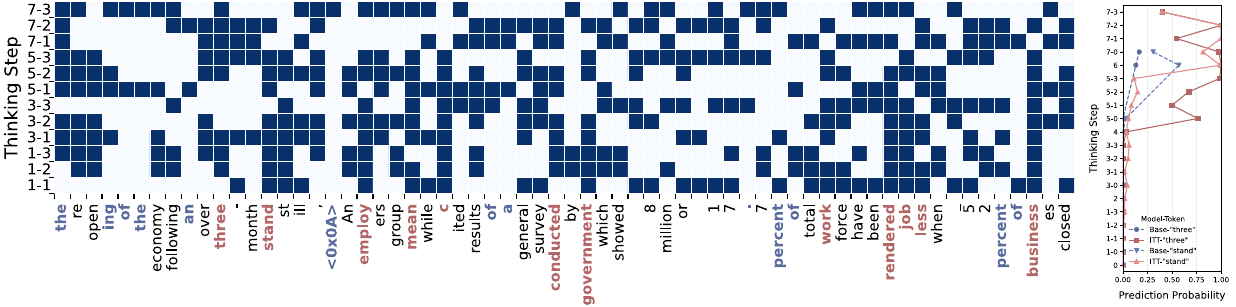 }
  \caption{ \textbf{Left:} Visualization of inner thinking routers' choices in \aname x4 -162M. "3-2" refers to the second thinking step in the 3rd layer (\aname layer). \aname allocates slow thinking to \textcolor{red!60!black}{difficult tokens} and fast thinking to \textcolor{blue!60!black}{easy tokens}. \textbf{Right:} The prediction probabilities for the tokens 'three' and 'stand' from \textcolor{blue!60!black}{LLaMA} and \textcolor{red!60!black}{ITT}.}\label{fig:router_visual}
\end{minipage}
\vspace{-2mm}
\end{figure*}

\paragraph{Adaptive Token Routing.} Disabling the dynamic routing mechanism results in a moderate PPL. increase (+0.19), but significantly impacts computational efficiency. This demonstrates the router's dual role: while maintaining prediction quality through selective processing, it achieves more than 50\% FLOPs reduction by focusing computation on 50\% most critical tokens in each step.

\paragraph{Router Setting.} Our experiments validate three critical design choices: The RTC design of ITT relies on explicit token selection signals (e.g., a 0.5 threshold in Sigmoid) for error correction and progressive disambiguation. The cumulative probability characteristic of Top-P conflicts with this deterministic routing mechanism, leading to a disruption in the iterative processing chain of key tokens. Sigmoid Normalization outperforms Tanh by 0.13 PPL., as it provides unambiguous activation signals for token selection whereas Tanh's negative values may disrupt RTC. Only Select Reweighting surpasses symmetric approaches by 0.15 PPL. through focused computation – selectively enhancing critical tokens while preserving original features for others. This targeted refinement minimizes interference between primary and augmented features.

\subsection{Analysis} \label{sec.exp.ana}

\paragraph{More Thinking for Better Performance.} As shown in Figure~\ref{fig:res_3figs} Left, \textbf{the performance gains from \aname's deep thinking mechanism do not diminish with more iterations}, unlike Loop's diminishing returns. The 162M \aname $\times$4 configuration improves 0.6\% over $\times$3, while Loop $\times$4 only shows a 0.3\% gain over $\times$3. This suggests that with sufficient computational resources, increasing \aname's thinking steps can unlock additional capabilities. The architectural advantage of \textbf{\aname becomes more apparent with larger model widths}, implying that smaller ITT models can adopt wider hidden dimensions to boost representational capacity.

\paragraph{Deeper Thinking with Fewer Tokens.} In Table~\ref{tab:elastic}, \aname x4 reduces the selection rate of the 4th step to 50\%, achieving a -0.26 PPL reduction compared to the training config, showing that \textbf{fewer tokens are needed for deeper thinking steps.} Additionally, \textbf{different thinking steps compensate for each other}, maintaining a PPL advantage of over 0.7 even when a step is removed. Figure~\ref{fig:res_3figs} Middle shows the average Position Encoding values, indicating that the model prioritizes earlier steps while assigning high weights to deeper ones. This demonstrates the \textbf{model's ability to optimize deep thinking with fewer, more impactful tokens}, with potential for even deeper thinking steps.

\paragraph{Routing Analysis.} Visualization of token selection paths (Figure~\ref{fig:router_visual}) demonstrates that approximately 30\%-50\% of tokens receive iterative thinking, with task-critical tokens (e.g., verbs, semantic pivots in red) being more likely to undergo multi-step thinking than low-information tokens. Moreover,the dynamic routing exhibits complementary thinking across steps:  In consecutive steps, \textbf{important tokens are prioritized for deeper thinking}. However, the 3-3 and 7-3 steps demonstrate compensatory choices for broader thinking. These two steps focus on simple tokens that were not given attention in previous steps, compensating for any missed details. Finally, interpretability analysis in Figure~\ref{fig:router_visual} Right demonstrate that\textbf{ ITT extend inner thinking steps, thereby preventing the failures observed in the baseline model.} This routing strategy developed during training, allows \textbf{ITT to achieve both depth and comprehensiveness}.

\section{Related Work}

\paragraph{Recurrent Computation} The concept of recurrence in machine learning traces back to foundational works on neural computation \citep{braitenberg1986vehicles} and LSTM networks \citep{gers2000lstm}. Modern extensions integrate recurrence into transformers through depth recurrence \citep{dehghani2019universal,lan2019albert,ng2024loopneuralnetworksparameter}. Recent works have re-discovered this idea for implicit reasoning~\cite{deng2023implicit,hao2024traininglargelanguagemodels} and test-time scaling~\cite{geiping2025scalingtesttimecomputelatent}. In contrast, \aname establishes a general-purpose recursive reasoning framework within individual layers and designs the Residual Thinking Cnnection (RTC) for enhanced capability.

\paragraph{Dynamic Computation Allocation} Dynamic Computation Allocation, like Mixture-of-Expert (MoE), reduce computational overhead by activating only a subset of networks~\cite{fedus2022switch, riquelme2021scaling, zhou2022mixture, jiang2024mixtral, xue2024openmoe}. Some works focus on elastic computation in depth, such as early exit~\cite{elhoushi2024layerskip,chen2023ee_llm}, parameter sharing~\cite{li2024lisa, li2024crosslayer} or using token-routing for dynamic layer skipping~\cite{zhang2024pmodbuildingmixtureofdepthsmllms}. Inspired by these works, ITT designs an elastic deep thinking architecture with Adaptive Token Routing (ATR) for efficient and adaptive computational resources allocation.

\section{Conclusion}
We propose \aname, a dynamic architecture enabling LLMs to allocate additional computation to critical tokens through adaptive inner thinking steps. By integrating token-wise depth routing, residual thinking connections, and step encoding, \aname enhance inner thinking without parameters expansion. Experiments demonstrate its potential for balancing efficiency with enhanced capabilities. 


\section*{Limitations}

While \aname demonstrates promising results, several limitations warrant discussion: First, our current implementation employs fixed routing patterns during training, potentially limiting dynamic adaptation to diverse token complexities. Second, our experiments focus on models up to 466M parameters - validation at larger scales could reveal new architectural interactions. Third, the residual thinking connections introduce additional memory overhead during backward passes, requiring optimization for industrial deployment. Finally, while our step encoding effectively differentiates thinking stages, more sophisticated temporal modeling might further enhance reasoning depth. These limitations present valuable directions for future research.

\section*{Ethical Considerations}
Our work adheres to ethical AI principles through three key aspects: 1) All experiments use publicly available datasets with proper anonymization, 2) The enhanced parameter efficiency reduces environmental impact from model training/inference, and 3) Our architecture-agnostic approach promotes accessible performance improvements without proprietary dependencies. We acknowledge potential risks of enhanced reasoning capabilities being misapplied, and recommend implementing output verification mechanisms when deploying \aname-based systems. Our work is committed to advancing accessible and efficient NLP technologies, fostering a more inclusive and automated future for AI.

\section*{Acknowledgments}
We would like to thank members of the IIE KDsec group for their valuable feedback and discussions. We sincerely thank Sean McLeish for his diligent review and critical feedback on this work. We are very grateful to Mengzhou Xia for providing the concise and effective ShearingLLaMA experimental code and for her assistance during the reproduction process. Work done during Yilong Chen's internship in Baidu Inc. This research is supported by the Youth Innovation Promotion Association of CAS (Grant No.2021153).

\bibliography{custom}

\newpage
\appendix

\section{Appendix}

\subsection{Algorithm}
As described in Section~\ref{sec.method}, the core algorithm of our proposed Inner Thinking Transformer implements fine-grained token-level reasoning optimization through dynamic depth computation. The detailed procedure is presented in Algorithm~\ref{alg:l4resrep-posmodblockv1}, which features three key innovations: 

\begin{itemize}
    \item \textbf{Adaptive Capacity Scheduling} with temperature annealing: The $\text{getCapacity}$ function gradually increases processed token count during initial training stages, enabling coarse-to-fine learning dynamics.
    
    \item \textbf{Hierarchical Residual Architecture}: Each thinking step $t$ scales and fuses current results ($\alpha^{(t)}\cdot\phi^{(t)}$) with positional encoding before integrating with previous hidden states.
    
    \item \textbf{Multi-grained Routing Network} utilizes hierarchical routing modules $\{\mathcal{R}^{(0)},...,\mathcal{R}^{(T)}\}$ to automatically identify critical tokens at different depth levels.
\end{itemize}

Notably, when training step $P$ stabilizes, the processing capacity $C$ progressively expands to cover all tokens, equipping the network with self-adaptive depth allocation capabilities. Theoretically, this algorithm extends the model's effective depth to $T+1$ times the baseline while maintaining FLOPs overhead of merely $O(kT/S)$. This establishes a parameter-efficient approach for enhancing reasoning capacity through explicit computation budgeting.

\subsection{Extend Related Work}

\paragraph{Recurrent Computation}
The concept of recurrence in machine learning traces back to foundational works on neural computation \citep{braitenberg1986vehicles} and LSTM networks \citep{gers2000lstm}. Modern extensions integrate recurrence into transformers through depth recurrence \citep{dehghani2019universal,lan2019albert,ng2024loopneuralnetworksparameter}, with recent improvements demonstrating algorithmic generalization via randomized unrolling \citep{schwarzschild2021randomized,mcleish2024}. From an optimization perspective, these models relate to energy-based gradient dynamics \citep{lecun2006contrastive} and test-time adaptation \citep{boudiaf2022testtime}. Recent works have introduced it for implicit reasoning~\cite{deng2023implicit,hao2024traininglargelanguagemodels} and test-time scaling~\cite{geiping2025scalingtesttimecomputelatent}. Inspired by these, ITT focuses on recursive reasoning within individual layers and designs the RTC architecture with theoretical support to enhance this capability.

\paragraph{Dynamic Computation Allocation}Dynamic Computation Allocation in architectures, like Sparse Mixture-of-Expert (MoE), utilize input adaptivity to reduce computational overhead by activating only a subset of subnetworks, or "experts," for each input token \cite{fedus2022switch, riquelme2021scaling, zhou2022mixture, jiang2024mixtral, xue2024openmoe}. Recent developments have introduced heterogeneous experts, integrating experts with varying capacities and specializations \cite{wu2024multihead, he2024millionexperts, dean2021pathways, zhou2022mixture}. Some works focus on elastic computation in depth, such as early exit~\cite{elhoushi2024layerskip,chen2023ee_llm}, parameter sharing~\cite{li2024lisa, li2024crosslayer} or using token-routing for dynamic layer skipping (Mixture of Depth)~\cite{zhang2024pmodbuildingmixtureofdepthsmllms}. Inspired by these works, ITT designs an elastic deep thinking architecture and uses Residual Thinking Connections to address the issue of non-continuous layer skipping.

\begin{algorithm*}[h]
\small
\caption{Forward Pass of the Inner Thinking Block}
\label{alg:l4resrep-posmodblockv1}
\begin{algorithmic}[1]
\Require 
    \textbf{Input:} Input tensor: $\mathbf{x} \in \mathbb{R}^{B \times S \times D}$, 
    Past key-value: $KV_{\text{past}}$, Attention mask: $\mathbf{M}$,
    Model parameters: $\Theta$, thinking steps $T$, training steps $P$, select rate $\rho$, warm-up steps $\tau$ 
\Ensure 
    $\mathbf{y} \in \mathbb{R}^{B \times S \times D}$, $KV_{\text{new}}$ \Comment{Output tensor and updated key-values}

\Statex Initialization: Routers: $\mathcal{R} = \{\mathcal{R}^{(0)}, \dots, \mathcal{R}^{(T)}\}$, 
Position weights: $\boldsymbol{\phi} = \{\phi^{(0)}, \dots, \phi^{(T)}\}$,  Scaling : $\boldsymbol{\alpha} = \{\alpha^{(0)},\dots, \alpha^{(T)}\}$ 

\State $\mathbf{y}^{(0)\prime}, KV_{\text{new}} \gets f(\mathbf{x}, KV_{\text{past}}, \mathbf{A}, \mathbf{M}, \Theta)$, \quad  $\mathbf{y}^{(0)} \gets \mathbf{y}^{(0)\prime} \odot \phi^{(0)}$ \Comment{Perform initial forward pass}
\State  $C \gets \text{getCapacity}(P, \rho, \tau)$ , \quad $k \gets \max(1, \lfloor C \cdot S \rfloor)$ \Comment{Compute routing weights, capacity}
\State  $\mathbf{W}^{(0)} \gets \mathcal{R}^{(0)}(\mathbf{y})$, \quad  $\mathcal{M}^{(0)} \gets \text{TopK}(\mathbf{W}^{(0)}, k)$ \Comment{Select top-$k$ tokens}

\For{$l = 1$ to $T$} \Comment{Iterate over maximum steps}

    
    
    \Statex \quad $\mathbf{y}_{\mathcal{M}^{(t-1)}}^{(t)\prime}, KV_{\text{new}} \gets f(\mathbf{y}^{(t-1)}_{\mathcal{M}^{(t-1)}}, KV_{\text{new}},  \mathbf{A}, \mathbf{M}, \Theta)$ \Comment{Perform selective forward pass}
    
    \Statex \quad $\mathbf{y}^{(t)} \gets \mathbf{y}^{(t-1)} + (\mathbf{y}^{(t-1)}_{\overline{\mathcal{M}^{(t-1)}}} +\alpha^{(t)}\cdot \mathbf{y}^{(t)\prime}_{\mathcal{M}^{(t-1)}}) \odot \phi^{(t)}$ \Comment{Scale and add selective output}
     \Statex \quad $\mathbf{W}^{(t)} \gets \mathcal{R}^{(t)}(\mathbf{y})$, \quad $\mathcal{M}^{(t)} \gets \text{TopK}(\mathbf{W}^{(t)}, k)$ \Comment{Compute routing weights, capacity}
\EndFor

\State \Return $\mathbf{y}^{(t)}, KV_{\text{new}}$

\end{algorithmic}
\end{algorithm*}
\subsection{Theoretical Proof of Multi-Step Residual Thinking Connection's Convergence}

In this Section, we provide a theoretical derivation showing that multi-step residual learning, used in Transformer architectures, is more effective than direct one-step learning in terms of gradient flow and convergence. We show that the multi-step process allows for more stable gradient propagation and faster convergence through geometric decay of the error, in contrast to the difficulties caused by gradient vanishing or explosion in direct one-step learning.

In deep learning models, especially in transformer-based architectures, the issue of gradient propagation across multiple layers has been a key challenge. Residual learning, where each layer updates the model with small corrections rather than directly mapping inputs to outputs, has shown promise in improving the stability of training and facilitating deeper networks. In this section, we will theoretically compare multi-step residual learning with direct one-step mapping to highlight why the former leads to better convergence and stability.

Let us consider the overall goal of a Transformer model. The final output \( F(x; \Theta) \) is a function of the input \( x \), parameterized by the model's parameters \( \Theta \), and is trained to minimize the loss function
\[
\mathcal{L}(F(x; \Theta), y^*)\,,
\]
where \( y^* \) is the target output.

For a single block \( B \) within the Transformer, we define an iterative process where the output at step \( k \), denoted by \( y_k \), is updated by adding a small residual term:
\[
y_{k+1} = y_k + \Delta_k(y_k; \theta)\,,
\]
where \( \theta \) is the shared parameter used for the residual function \( \Delta_k \). The goal is to iteratively refine the output by accumulating these residuals. After \( K \) iterations, the final output becomes:
\[
y_K = y_0 + \sum_{k=0}^{K-1} \Delta_k(y_k; \theta)\,,
\]
where \( y_0 \) is the initial input to the block.

\paragraph{Gradient Propagation in Direct One-Step Mapping}
In the direct one-step mapping, we try to learn the function \( F(x; \theta) \) directly from the input to the output. The loss function is defined as:
\[
\mathcal{L} = \mathcal{L}(F(x; \theta), y^*)\,.
\]
The gradient of the loss function with respect to the parameters \( \theta \) is:
\[
\frac{\partial \mathcal{L}}{\partial \theta} = \frac{\partial \mathcal{L}}{\partial F(x; \theta)} \cdot \frac{\partial F(x; \theta)}{\partial \theta}\,.
\]
In deep networks, the term \( \frac{\partial F(x; \theta)}{\partial \theta} \) involves multiple layers of non-linear transformations. This can cause the gradients to either vanish or explode as they propagate back through the layers, leading to unstable training. Specifically, when \( \theta \) is deep within the network, the gradient may be subject to shrinking (vanishing) or growing (exploding) due to the repeated chain rule applications, which impedes effective training.

\paragraph{Gradient Propagation in Multi-Step Residual Learning}
Now, we consider the multi-step residual learning process. After \( K \) iterations, the output of the block is:
\[
y_K = y_0 + \sum_{k=0}^{K-1} \Delta_k(y_k; \theta)\,.
\]
We want to compute the gradient of the loss function \( \mathcal{L} \) with respect to the shared parameters \( \theta \). Using the chain rule, the gradient of \( y_K \) with respect to \( \theta \) is:
\[
\frac{\partial y_K}{\partial \theta} = \frac{\partial y_K}{\partial y_{K-1}} \cdot \frac{\partial y_{K-1}}{\partial y_{K-2}} \cdots \frac{\partial y_1}{\partial \theta}\,.
\]
For each residual update, we have:
\[
\frac{\partial y_{k+1}}{\partial y_k} = I + \frac{\partial \Delta_k(y_k; \theta)}{\partial y_k}\,,
\]
where \( I \) is the identity matrix, and \( \frac{\partial \Delta_k(y_k; \theta)}{\partial y_k} \) represents the gradient of the residual function. Therefore, the total gradient is:
\[
\frac{\partial y_K}{\partial \theta} = \prod_{k=0}^{K-1} \left(I + \frac{\partial \Delta_k(y_k; \theta)}{\partial y_k}\right) \cdot \frac{\partial \Delta_0(y_0; \theta)}{\partial \theta}\,.
\]
If each residual update \( \Delta_k(y_k; \theta) \) is small, we can approximate:
\[
I + \frac{\partial \Delta_k(y_k; \theta)}{\partial y_k} \approx I\,.
\]
This leads to:
\[
\frac{\partial y_K}{\partial \theta} \approx \frac{\partial \Delta_0(y_0; \theta)}{\partial \theta}\,.
\]
Thus, the gradient flow in each step is relatively stable and doesn't suffer from drastic shrinking or explosion, allowing for efficient and stable training.

\paragraph{Convergence in Direct One-Step Learning}
For direct one-step learning, the model learns the entire transformation from \( x \) to \( y \) in one step, which can be represented as:
\[
y = F(x; \theta)\,.
\]
The training objective is to minimize the loss function:
\[
\mathcal{L} = \mathcal{L}(F(x; \theta), y^*)\,.
\]
However, due to the complexity of the non-linear function \( F(x; \theta) \), the gradients can either vanish or explode as they propagate through the layers. In the worst case, the gradients may become extremely small (vanishing gradients) or extremely large (exploding gradients), causing the optimization process to stall or fail to converge to an optimal solution.

\paragraph{Convergence in Multi-Step Residual Learning}
In multi-step residual learning, each step updates the output with a small correction, and the final output is the sum of all the incremental corrections. The error at step \( k \) is given by:
\[
e_k = T(x) - y_k\,,
\]
where \( T(x) \) is the target. The error at step \( k+1 \) is:
\[
e_{k+1} = T(x) - y_{k+1} = e_k - \Delta_k(y_k; \theta)\,.
\]
If the residual updates \( \Delta_k(y_k; \theta) \) are small, the error at each step decreases geometrically:
\[
\| e_{k+1} \| \leq c \| e_k \| \quad \text{for some constant} \quad 0 < c < 1\,.
\]
After \( K \) iterations, the error will decrease exponentially:
\[
\| e_K \| \leq c^K \| e_0 \|\,.
\]
This shows that the error decays exponentially with the number of steps, leading to fast convergence as the number of iterations increases.

\subsection{Extend Analysis} 

\paragraph{Router Weights Visulization} The observed normal distribution of routing weights in the ITT framework, with its distinctive concentration within the 0.6-0.8 range, emerges as a self-regulating mechanism that fundamentally reconciles computational efficiency with model effectiveness. This central tendency facilitates dynamic resource allocation through probabilistic token selection, where moderately high weights enable smooth computational load balancing while preserving residual information pathways. The distribution's avoidance of extreme values inherently supports flexible top-k adjustments, allowing the system to scale computation across contexts without abrupt performance degradation - a critical feature for processing variable-length inputs and maintaining throughput consistency.

The weight concentration further ensures training stability through continuous differentiability across routing decisions. By preventing abrupt 0/1 selection thresholds, the architecture maintains stable gradient flows during backpropagation, effectively distributing learning signals between activated and bypassed tokens. 

\label{sec:appendix}

\begin{table}[t]
\centering
\small
\setlength{\tabcolsep}{2.5mm}{
\begin{tabular}{@{}lcc@{}}
\toprule
\textbf{Method - Select Ratio in Steps} & \textbf{FLOPs} & \textbf{Perplexity} $\downarrow$\\
\midrule
LLaMA2-162M & 1.88 & 11.13 \\
\midrule
\aname×4 - 90\%,~90\%,~90\% & 4.42 & 10.27 (\textcolor{green!70!black}{-0.86}) \\
\midrule
\aname×4 - 90\%,~90\%,~0\% & 3.57 & 10.40 (\textcolor{green!70!black}{-0.73}) \\
\aname×4 - 90\%,~0\%,~90\% & 3.57 & 10.36 (\textcolor{green!70!black}{-0.77}) \\
\aname×4 - 0\%,~90\%,~90\% & 3.57 & 10.56 (\textcolor{green!70!black}{-0.57}) \\
\midrule
\aname×4 - 90\%,~90\%,~70\% & 4.23 & 10.25 (\textcolor{green!70!black}{-0.88}) \\
\aname×4 - 90\%,~70\%,~90\% & 4.23 & 10.23 (\textcolor{green!70!black}{-0.90}) \\
\aname×4 - 70\%,~70\%,~90\% & 4.04 & 10.21 (\textcolor{green!70!black}{-0.92}) \\
\aname×4 - 90\%,~70\%,~70\% & 4.04 & 10.22 (\textcolor{green!70!black}{-0.91}) \\
\midrule
\aname×4 - 70\%,~70\%,~70\%$^\dagger$  & 3.85 & 10.52 (\textcolor{green!70!black}{-0.61}) \\
\midrule
\aname×4 - 70\%,~70\%,~50\% & 3.66 & 10.26 (\textcolor{green!70!black}{-0.87}) \\
\aname×4 - 70\%,~50\%,~70\% & 3.66 & 10.26 (\textcolor{green!70!black}{-0.87}) \\
\aname×4 - 50\%,~70\%,~70\% & 3.66 & 10.29 (\textcolor{green!70!black}{-0.84}) \\
\midrule
\aname×4 - 70\%,~50\%,~50\% & 3.47 & 10.34 (\textcolor{green!70!black}{-0.79}) \\
\aname×4 - 50\%,~50\%,~70\% & 3.47 & 10.36 (\textcolor{green!70!black}{-0.77}) \\
\aname×4 - 50\%,~70\%,~50\% & 3.47 & 10.34 (\textcolor{green!70!black}{-0.79}) \\
\aname×4 - 50\%,~50\%,~50\% & 3.29 & 10.47 (\textcolor{green!70!black}{-0.66}) \\
\midrule
Loop×4 - 100\%,~100\%,~100\%$^\dagger$  & 4.70 & 10.78 (\textcolor{green!70!black}{-0.35}) \\
\bottomrule
\end{tabular}
}

\caption{Eval Perplexity in the \aname setting is performed for extend 3 steps' thinking. $^\dagger$ refers to the model's training configuration.}\label{tab:elastic}
\vspace{-2mm}
\end{table}

\begin{table}[t]
\centering

\small{\resizebox{0.9\columnwidth}{!}{%
\begin{tabular}{@{}lcccc@{}}
\toprule
\textbf{Model Setting} & \textbf{L.2-162M} &  \textbf{L.2-230M} &   \textbf{L.2-466M} \\ \midrule
\textit{hidden size }        & 1024  & 1536 & 2048  \\
\textit{intermediate size }        &  2560  & 2560 & 4096  \\
\textit{attention heads}         &  32  & 32 & 32  \\
\textit{num kv heads}      &  32  & 16 & 32 \\
\textit{layers }         & 8  & 8 & 8\\
\midrule
\textbf{\# Params}  & 162M   & 230M & 466M \\
\bottomrule
\end{tabular}
}}
\caption{Detailed configuration, activation parameters, and total parameters of the models included in our study. L.2-162M represents the LLaMA-2 architecture model with 162M total parameters.}
\label{tab:model_setting}
\end{table}

\end{document}